\documentclass[sigconf, authorversion]{acmart}
\usepackage[utf8]{inputenc}
\usepackage{amsmath}
\usepackage{graphicx}

\usepackage{xcolor}
\usepackage{float}
\usepackage{fancyvrb}
\usepackage{listings}
\begin{document}

\title[Narrative Analysis with KGLLMs]{Narrative Analysis of True Crime Podcasts With Knowledge Graph-Augmented Large Language Models}

\author{Xinyi Leng}
\affiliation{
  \institution{Carleton College, Department of Mathematics and Statistics}
  \country{}
  }

\author{Jason Liang}
\affiliation{
  \institution{Claremont McKenna College, Department of Mathematical Sciences}
  \country{}
  }

\author{Jack Mauro}
\author{Xu Wang}
\author{James Chapman}
\author{Andrea L. Bertozzi}
\affiliation{
  \institution{University of California, Los Angeles, Department of Mathematics}
  \country{}
  }

\author{Junyuan Lin}
\affiliation{
  \institution{Loyola Marymount University, Department of Mathematics, Statistics and Data Science}
  \country{}
  }

\author{Bohan Chen}
\affiliation{
  \institution{California Institute of Technology, Computing + Mathematical Sciences, Division of Engineering and Applied Science}
  \country{}
  }

\author{Chenchen Ye}
\affiliation{
    \institution{University of California, Los Angeles, Department of Computer Science}
    \country{}
    }

\author{Temple Daniel}
\author{P. Jeffrey Brantingham}
\affiliation{
  \institution{University of California, Los Angeles, Department of Anthropology}
  \country{}
  }

\renewcommand{\shortauthors}{Leng, et al.}

\begin{abstract}
    Narrative data spans all disciplines and provides a coherent model of the world to the reader or viewer. Recent advancement in machine learning and Large Language Models (LLMs) have enable great strides in analyzing natural language. However, Large language models (LLMs) still struggle with complex narrative arcs as well as narratives containing conflicting information. Recent work indicates LLMs augmented with external knowledge bases can improve the accuracy and interpretability of the resulting models. In this work, we analyze the effectiveness of applying knowledge graphs (KGs) in understanding true-crime podcast data from both classical Natural Language Processing (NLP) and LLM approaches. We directly compare KG-augmented LLMs (KGLLMs) with classical methods for KG construction, topic modeling, and sentiment analysis. Additionally, the KGLLM allows us to query the knowledge base in natural language and test its ability to factually answer questions. We examine the robustness of the model to adversarial prompting in order to test the model's ability to deal with conflicting information. Finally, we apply classical methods to understand more subtle aspects of the text such as the use of hearsay and sentiment in narrative construction and propose future directions. Our results indicate that KGLLMs outperform LLMs on a variety of metrics, are more robust to adversarial prompts, and are more capable of summarizing the text into topics. 
\end{abstract}



\received{16 August 2024}
\received[revised]{6 September 2024} 
\received[accepted]{30 August 2024}

\acmConference[GTA3 Workshop-2024]{ACM Conference}{October 2024}{33rd International Conference on Information and Knowledge Management, Boise, Idaho, USA}
\acmDOI{}
\maketitle

\section{Introduction}
Knowledge graphs are powerful tools for organizing, storing, and presenting data with complex relationships between diverse types of objects, such as text data from various online platforms, and have been successfully applied to a wide range of data-driven research problems~\cite{flocco2021analysiscovid19knowledgegraph,adams2022knowledge,bozhidarova2023hate}. A knowledge graph is a graph whose nodes represent entities of interest and whose edges represent relationships between those entities \cite{wang2017knowledge, ji2021survey}. This work seeks to use knowledge graphs to analyze the true crime genre, specifically the podcast \textit{Serial} \cite{koenig2014serial}. The true crime genre poses interesting challenges for knowledge graph generation as the data is organized in a narrative format rich with emotion, conflict and contradictory information. 

Large Language Models (LLMs) are machine learning models designed for processing text data. They are trained on massive amounts of data (typically large fractions of the internet) and thus encode an enormous amounts of information. The breadth of information in their training data allows them to excel at a wide variety of natural language tasks. More recent LLMs have demonstrated reasoning capabilities and surpass humans on a variety of benchmarks \cite{webb2023LLMs, katz2024gpt, street2024llms}.

However, LLMs still suffer from some key limitations. These models are known to hallucinate information which may be non-factual \cite{Ji_2023, huang2023surveyhallucinationlargelanguage}. They also struggle to process large amounts of text due to their limited context window. Recent works address this issue by augmenting the LLM with a mechanism for information storage and/or retrieval. This process, known as Retrieval Augmented Generation (RAG), allows the model to directly interact with stored information thus improving recall in long texts. It also tends to generate more factual responses than standard LLMs \cite{lewis2021retrievalaugmentedgenerationknowledgeintensivenlp}. There is significant interest in connecting RAG with structured data representations like with KGs \cite{chen2023autokg, han2024pive, Pan_2024}. Because knowledge graphs encode information about the relationships between entities in the data, they may be better equipped to capture the complex relationships between people and events in a narrative. A KGLLM seeks to combine KGs and LLMs by using an LLM in the KG construction process and/or using the KG in the LLM querying process. For a broader overview of KGLLMs, we refer the interested reader to the survey paper by Pan et al. \cite{Pan_2024}.

Previous studies have explored knowledge graphs applied to fictional crime narratives \cite{alaverdian2020killed}. This work was produced before LLMs were widely available and relies on a manually constructed KG. Tools like VADER for sentiment analysis and LDA for topic modeling have long been used to study narratives \cite{Hutto_Gilbert_2014, LDApaper}. More recently, researchers have investigated applications of LLM-based tools like BERT to narrative analysis \cite{jones2019exploration}.

In this work, we focus on GraphRAG which performs both knowledge graph construction and retrieval \cite{edge2024localglobalgraphrag}. GraphRAG automates the knowledge graph construction which may be biased by our understanding of the story and reduces the amount of human labor required. It also has the potential to add new sources of data as the case evolves. Additionally, it is designed to integrate directly with the KG and gives us a way to evaluate the impact of the KG via its performance on a variety of natural language tasks.

\subsection{The \textit{Serial} Podcast}
\label{sec:serial}

\textit{Serial: Season One} is a true crime podcast hosted by Sarah Koenig \cite{koenig2014serial}. Upon its premier in October 2014, \textit{Serial} quickly amassed millions of listeners and became a cultural phenomenon. The first season centers around the 1999 murder of Hae Min Lee and the conviction of her ex-boyfriend, Adnan Syed \cite{albrecht2023serial}. Both were high school students from Baltimore, Maryland. To this day, Adnan has maintained his innocence. Sarah guides listeners through an investigation of the circumstances surrounding Hae’s murder. Assisted by forensic experts, she interviews witnesses and examines police reports, news articles, and trial testimony. Though Sarah amasses numerous facts about the case, she faces contradictory accounts and unrecoverable evidence. The podcast implies that a definitive conclusion is impossible to reach. 

This dataset presents interesting challenges for machine learning applications. Principally, the dataset contains conflicting information and sources which may not be credible in a court of law. The intended audience of the podcast is the public and the information may be presented with some aspect of entertainment in mind. This deviates from similar work in applying KGLLMs where the dataset is consistent with itself and the data is generally trustworthy \cite{chen2023autokg, edge2024localglobalgraphrag}. Additionally, the order in which information is presented can influence the reception of facts. Notably, the podcast does not progress in chronological order, so the neural network needs to consider not only the order of events in the narrative, but also the chronological order in which they happen. LLMs are also susceptible to many human biases which may be present in a narrative that would likely not exist in other works \cite{acerbi2023large}. Lastly, the LLM will have to answer questions about a topic for which there are no definitive answers. It needs to deal with missing information, uncertainty about the outcomes, and information only implied by the text. It is of utmost importance to understand model hallucinations in this context to ensure factual accuracy. 

\subsection{Contributions}
We preprocess the original podcast audio data into a text format. Following this, we present the complexities of the narrative format and demonstrate shortcomings of classical analysis on the dataset. We propose KGLLMs to tackle the problem due to their ability to construct and encode relational information in the KG and their ability to reason about inputs and the data stored in the KG. Next we compare the results of the KGLLM (GraphRAG) to those of classical methods and LLMs. Additionally, we perform both sentiment analysis to understand the narrative format and analysis of hearsay statements to assess the legal credibility of information presented. Finally, we present future directions that we feel are important for understanding KGLLMs and their application to narrative data. Our code is made publicly available. \footnote{https://github.com/jmauro1/Narrative-Analysis-True-Crime-Podcasts-KGLLM}

In this paper, we explore the integration of KGLLMs used to analyze the true crime genre, and more specifically, the \textit{Serial} podcast. After presenting background information relevant to our study in Section~\ref{sec:background}, we introduce the methodology used to analyze the data in Section~\ref{sec:methods} and discuss our experimental results in Section~\ref{sec:experiments}. Finally, we end with a discussion of the work and propose future directions in Section~\ref{sec:discussion}.

\section{Background}
\label{sec:background}
In this section, we provide an overview of the concepts used in our methodology and experiments. We present background information on knowledge graphs, and we highlight the main aspects of adversarial prompting, which is the main way we test robustness against hallucinations of the models used to analyze the narrative. Sections~\ref{sec:topicModelingBackground}-\ref{sec:hearsayBackground} provide an overview of standard narrative analysis techniques.

\subsection{Knowledge Graphs}
\label{sec:KGBackground}
A knowledge graph $G = \left( \mathcal{E}, \mathcal{R}\right)$ is a digraph which represents facts in a structured and searchable manner. The entity set $\mathcal{E}$ is the set of nodes and their corresponding labels. The relation set $\mathcal{R}$ is the set of ordered triples $(h, r, t)$ where $r$ denotes the relationship between entities $h$ and $t$. Note that this relationship is often directional: (Jane, called, the post office). 

A hierarchical knowledge graph $G = \left( \mathcal{E}, \mathcal{R}, \mathcal{C} \right)$ is a knowledge graph with a set $\mathcal{C}$ that encodes a hierarchical clustering of the nodes in the graph \cite{edge2024localglobalgraphrag}. In this paper, we consider hierarchical partitions
$$\mathcal{C} = \{P_l\ |\ P_l\text{ partitions } \mathcal{E}, P_{l-1} \text{ is a refinement of } P_l, l = 1, ..., h\}.$$
In other words, each level of the hierarchy divides the nodes of the graph, with lower level partitions containing many smaller groups. This allows the hierarchical knowledge graph to encode clusters corresponding to specific information in lower levels of the hierarchy and broad information in higher levels of the hierarchy. 

\subsection{Adversarial Prompting} 
\label{sec:advPromptingBackground}
A language model hallucination occurs when a response is generated which is nonsensical or unfaithful to source content. The tendency for language models to hallucinate is a significant concern. Huang et al. define two types of hallucinations: faithfulness hallucinations and factuality hallucinations \cite{huang2023surveyhallucinationlargelanguage}. Faithfulness hallucinations occur when language models do not follow the prompt instructions or perform faulty logic. Factuality hallucinations occur when the language model incorrectly recalls information or fabricates information. We are primarily interested in understanding factuality hallucinations due to the sensitive and/or controversial content of the podcast data. This is particularly important if any information extracted from an LLM is used to make determinations about human outcomes. 

To test the model, we consider \textit{adversarial prompts}, which are prompts designed in an attempt to cause LLM hallucinations \cite{yao2023llm}. Adversarial prompts can be used to assess the robustness of a LLM-assisted model by testing the degree to which it hallucinates. By varying the type and difficulty of the adversarial prompts, we may gain a finer understanding of model shortcomings. LLMs integrated with external knowledge bases may be less prone to factual hallucinations, especially when applied to questions involving information directly retrievable from the knowledge base. However, it is still important to test the model's ability to combine disparate information from the narrative as well as its ability to assess if the information in the knowledge base is factual. The latter point is particularly important when viewpoints conflict and pieces of evidence have different levels of credibility.

\subsection{Topic Modeling}
\label{sec:topicModelingBackground}
Topic modeling involves summarizing the text into a smaller collection of topics. This is important for higher level reasoning about the narrative and gives us an avenue to analyze model outputs \cite{hahn2000challenges}. Because the podcast data comprises many people, events, and pieces of evidence, it is important to assess the model's ability to both construct topics and understand their significance in the broader narrative. LLMs have the potential to construct more nuanced and relevant topics. We empirically compare topics generated by GraphRAG with those generated via LLMs. 

\subsection{Sentiment Analysis and Change-point Detection}
\label{sec:sentAnalysisBackground}
It is well known that the emotional presentation of information can influence the way that humans perceive knowledge and events. This phenomenon is also present in LLMs due to human biases present in the training data \cite{acerbi2023large}. The \textit{Serial} podcast is a true crime investigation of the events and evidence surrounding a murder. Since the audience of the podcast is the public, it may deviate from legal standards and employ narrative tactics to improve its popularity amongst a general audience. We refer to this narrative presentation as the entertainment narrative. Sentiment analysis may shed light on aspects of the entertainment narrative including how facts are presented and their role in constructing the narrative. Sentiment analysis involves classifying text as positive, negative, or neutral using various Natural Language Processing (NLP) methods, including rule-based, automatic, and hybrid approaches \cite{devika2016sentiment}. In this work, we compare classical sentiment analysis models and LLMs for understanding the impact of sentiment on the text. Our work primarily focuses on VADER and LLM analysis of sentiment \cite{Hutto_Gilbert_2014}.

\subsection{Hearsay}
\label{sec:hearsayBackground}
When investigating the podcast, it is important to assess the credibility of information presented in the text. Hearsay presents a complex legal challenge. According to Rule 801(c) \cite{d5a6e99d-5d92-3d7f-98c8-88f1ada88412}, hearsay is defined as a “statement” made outside of court, other than one made by the declarant while testifying, and offered to prove the truth of the matter asserted. The issue is that establishing credibility is challenging when the person being quoted is not present. Under Rule 802, hearsay is generally inadmissible unless an exception is provided by the Federal Rules of Evidence \cite{d5a6e99d-5d92-3d7f-98c8-88f1ada88412}. In contrast, podcasts like \textit{Serial} are not bound by such legal constraints. We investigate the prevalence and impact of hearsay in \textit{Serial} and how it contributes to the entertainment narrative and overall sentiment of the story.

\section{Methods}
\label{sec:methods}
In this section, we provide an overview of the preprocessing pipeline for our data. We also discuss the shortcomings of a traditional knowledge graph for this problem and motivate the need for an LLM-enhanced knowledge graph construction.

\subsection{Preprocessing}
Our dataset consists of the entire transcript from the \textit{Serial} Podcast Season 1 converted from audio to a text file using Whisper and further processed into a csv file \cite{radford2023robust}. This raw dataset has 2,055 rows in total, and each row of the dataset contains the following features: a unique sequence number, the episode number and title, the start and end times of the text segment, the raw text segment, and the speaker of the segment. The length of each text segment in the raw data varies, as each segment comes from a timestamp window that varies in length between only a couple seconds and roughly thirty seconds. 

To resolve ambiguity introduced by first person recounts of events, we replace the pronoun ``I'' with the speaker's name. Further, we remove the clitic from certain contractions ('ve from would've, 'd from I'd, etc.) without changing the meaning of the text and expand negative contractions to their full forms. To maintain the integrity of names and specific phrases, spaces between first names and surnames and spaces within proper nouns are removed. For example, `Adnan Syed' is changed to `AdnanSyed' and `Best Buy' is changed to `BestBuy.' 
Additionally, we label each text segment with its episode number and ordered timestamp. For example, a segment labeled `2\_51' means that segment is the 51st time segment in episode 2. Furthermore, we filter out irrelevant texts, such as repeated openings in each episode. After this procedure, the processed dataset contains 2,003 rows.

In addition to the above preprocessing techniques, an optional step involves removing discourse markers and filler words to make the text more concise. This optional step removes 148 rows from the dataset. This step is applied only for topic extraction using BERTopic in Section~\ref{sec:topicModeling}.

\subsection{KG Generation Methods}
In order to augment the narrative analysis process with an external knowledge base, we construct a knowledge graph over the preprocessed \textit{Serial} data. We begin by constructing a traditional knowledge graph. Our exploration of the graph revealed shortcomings of this approach, so we use GraphRAG \cite{edge2024localglobalgraphrag} for an LLM enhanced knowledge graph construction pipeline. 
\subsubsection{Method 1 (Traditional Knowledge Graph)} 

The Open Domain Information Extraction algorithm introduced in Angeli et al. is used to extract the relations $(h, r, t)$ in the traditional knowledge graph \cite{angeli-etal-2015-leveraging}. In Angeli et al., longer utterances are reduced to a set of self-contained clauses that maintain the original information conveyed in the utterance. This is done by training a multinomial logistic regression model to traverse the dependency parse tree of the utterance. Natural logic rules are then used to shorten the clauses to more compact segments. The triples $t$ are then extracted from these segments. Each triple $(h, r, t)$ is given a weight attribute $w \in \mathbb{R}$ where $w$ is the number of times that relation appears in the podcast. These weighted triples are then used to build the traditional knowledge graph.

The knowledge graph constructed from this model has 2,720 nodes and 6,640 edges. Analysis of this graph reveals that the graph contains details about pertinent information to the case. The graph captures important temporal information in the murder case, such as the day Hae Min Lee went missing or the timeline of Hae and Adnan's breakup. The graph also contains information about key evidence in the case. Hae's personal diary was admitted as evidence in Adnan's trial, and the graph includes information about the days Hae wrote in her personal diary and the content of these diary entries. Beyond the case information, the graph also expresses personal information about Hae, such as her interest in field hockey, love of the movie \textit{Titanic}, and her personality traits such as ``unshy'' and ``cheerful.'' 

\begin{figure}[htbp]
    \centering
    \includegraphics[width=.75\linewidth]{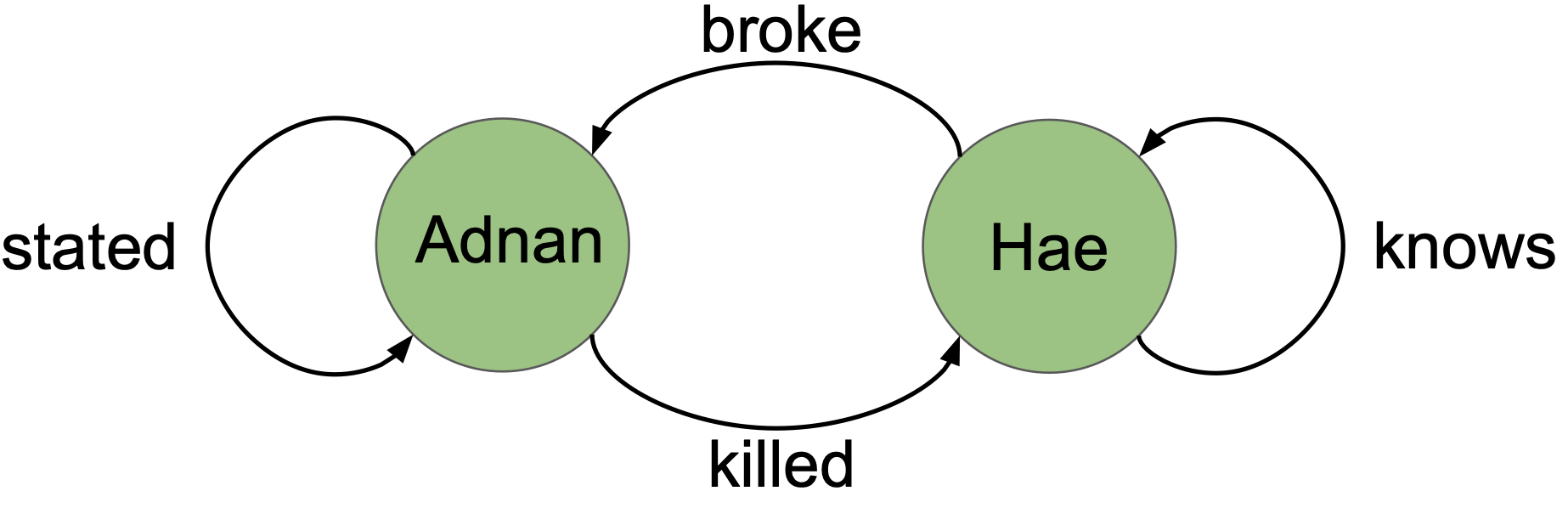}
    \caption{Subgraph Showing Limitations of Traditional KGs}
    \label{fig:adnanKilledHae}
    \Description[Traditional KG Subgraph]{Described in text.}
\end{figure}
While the traditional KG captures information about the case, it has important limitations. Many of the triples extracted include pronouns, diluting the utility of the information conveyed. Similarly, some relations and nodes in the graph are ambiguous in meaning, limiting the factual accuracy of the information provided in the triples. The inclusion of pronouns and ambiguous nodes and relations also leads to a high number of nodes and edges. Perhaps the most severe shortcoming is that the narrative structure of the data degrades factual accuracy of the extracted triples. Figure \ref{fig:adnanKilledHae} depicts a relation of "killed" extracted between the nodes of the accused Adnan Syed and victim Hae Min Lee. As the podcast serves to investigate Adnan's innocence, a claim such as this should not appear as ground truth in the knowledge graph. 

\subsubsection{Method 2 (GraphRAG)} 
GraphRAG \cite{edge2024localglobalgraphrag} is an automated way to build a knowledge graph, using an LLM to extract relations from the text. The knowledge graph constructed  is a hierarchical KG. GraphRAG takes input documents and uses an LLM to construct the entity set $ \mathcal{E}$ and relation set $ \mathcal{R}$. GraphRAG performs the entity and relation extraction process several times through a gleaning process that both prune out unnecessary nodes and relations and maximizes the amount of information captured. Once these nodes and relations are extracted, the Leiden algorithm \cite{Traag_2019} clusters the graph into hierarchical communities, resulting in the final graph  $G = \left( \mathcal{E}, \mathcal{R}, \mathcal{C} \right)$.  Once the graph is constructed, a LLM creates a node summary report, a relation summary report, and community summary reports. These reports are then used as external information in the RAG process. GraphRAG implements both a local and global retrieval mechanism for querying the KG. Local retrieval is designed for specific question answering, whereas global retrieval is designed for more abstract reasoning about the text \cite{edge2024localglobalgraphrag}.

\begin{figure}[htbp]
    \centering
    \includegraphics[width=.7\linewidth]{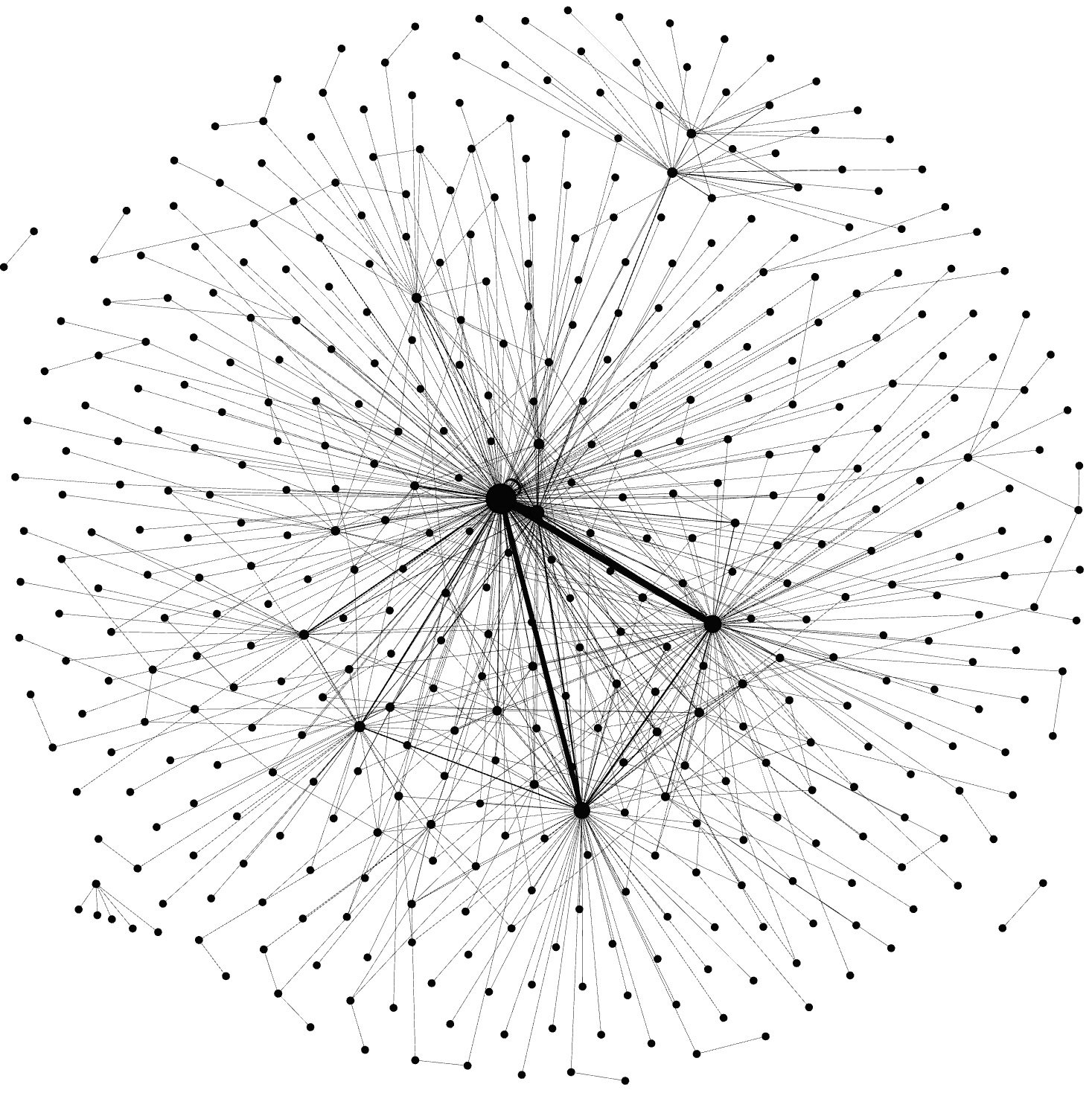}
    \caption{GraphRAG Knowledge Graph of the \textit{Serial} Podcast}
    \label{fig:graphRAGGraph}
    \Description[GraphRAG KG]{Described in text.}
\end{figure}

To construct a knowledge graph generated from the \textit{Serial} podcast data, we used the default parameters of GraphRAG with three modifications: the input chunk size was changed to 600 tokens, the LLM used throughout the process was changed to GPT-4o-mini, and the embedding model was changed to Open AI's text-embedding-3-small. The chunk size was adjusted because \cite{edge2024localglobalgraphrag} suggests 600 is an optimal chunk size for entity extraction and the base models were chosen to reduce experimentation costs. The graph constructed from this process has 462 nodes and 751 edges (see Figure~\ref{fig:graphRAGGraph}).
The largest nodes in this graph correspond to Adnan Syed, Hae Min Lee,  Jay Wilds, and Sarah Koenig. The edges are weighted based on normalized counts of detected information contributing to the relation the edge represents. We note that the edges in this graph contain paragraphs of text which allows complex relationships between entities to be captured. 

We find that the graph generated by GraphRAG overcomes the obstacles faced by traditional KGs. Pronouns and ambiguous nodes and relations have been pruned out of the graph. Furthermore, the information-rich edges allow for a distinction to be made between true and speculative claims, which is crucial for narrative data such as the \textit{Serial} podcast. 

\section{Experiments}
\label{sec:experiments}
In this section, we provide an overview of our experiments. Section~\ref{sec:queryResults} discusses query results using the GraphRAG model, and section~\ref{sec:adversarial_exp} focuses on query results from adversarial prompts. Sections~\ref{sec:topicModeling}-\ref{sec:hearsay} provide an analysis of the narrative, with section~\ref{sec:topicModeling} discussing topic modeling, section~\ref{sec:sentAnalysis} providing sentiment analysis of the podcast, and section~\ref{sec:hearsay} performing hearsay analysis. 
Unless otherwise stated, the base model used in all experiments is GPT-4o-mini. 

\subsection{Query Results using GraphRAG} 
\label{sec:queryResults}
To investigate how a KGLLM responds to queries, we evaluate the responses of GraphRAG to various prompts. 
To understand how the LLMs perform on narrative tasks, we developed a corpus of 36 questions about \textit{Serial}. Although this is a relatively small number of evaluations, each question has to be tailored to the particular dataset and the overall narrative. To address diversity in queries, the questions ask about ground truths, overarching themes, and opinions about the podcast. Responses from GraphRAG with both local and global search, RAG, and standalone GPT are compared. Responses from these models are evaluated by a separate LLM called an \textit{evaluator}, as laid out in Edge et al. \cite{edge2024localglobalgraphrag}. The evaluator compares answers based on four metrics: comprehensiveness, directness, diversity, and empowerment. Comprehensiveness assesses the depth of information and whether all aspects of the question are addressed. Directness assesses the clarity, specificity, and ambiguity of the question. Diversity assesses the range of perspectives, ideas, and examples provided. Empowerment assesses how the response equips one to form well-informed opinions on the topic. 

The evaluator is then prompted to choose which model provided the best response to each question according to each metric. These results are shown in Table~\ref{tab:model_metrics}. Of the 144 total evaluations, GraphRAG performed best on 131 of the evaluations while RAG and Naive LLM scored highest on 5 and 8 respectively. GraphRAG local search alone was the best choice for 110 of the evaluations. A possible explanation for the superior performance of local search is the type of questions asked. More questions were about fact retrieval than overarching themes, and local search may be better suited for these types of queries. It is worth noting that local search initially performs the RAG process and then supplements the information retrieved with details from the graph, which could provide an explanation as to why it outperforms RAG. The evaluation results show that GraphRAG responses consistently outperform RAG and GPT responses on these metrics.

\begin{table}[t]
    \centering
    \caption{Question-Answer Results. Com=Comprehensiveness, Emp=Empowerment, Div=Diversity, Dir=Directness}
    \begin{tabular}{cccccc}
        \toprule
        Model & Com & Emp & Div & Dir & Total\\
        \midrule
        GraphRAG Local & 27 & 30 & 29 & 24 & 110 \\
        GraphRAG Global & 8 & 5 & 6 & 2 & 21 \\
        Naive LLM & 1 & 1 & 1 & 5 & 8\\
        Naive RAG & 0 & 0 & 0 & 5 & 5\\
        \bottomrule
    \end{tabular}
    \label{tab:model_metrics}
    \Description[GraphRAG Local outperforms other models]{GraphRAG Local wins most Question-Answer evaluations (approximately 76 percent) when compared against GraphRAG Global, Naive LLM, and Naive RAG.}
\end{table}

\subsection{Adversarial Prompting} 
\label{sec:adversarial_exp}
In order to test robustness, we prompt GraphRAG local, GraphRAG global, RAG, and GPT with an additional 6 adversarial prompts. The prompts test for both factual fabrications and inconsistencies. These responses were manually evaluated because the correct answers are known. In general, GraphRAG demonstrates impressive resistance to inconsistencies while both RAG and GPT provide inconsistent responses on multiple occasions. For example, when the prompt included falsified evidence of Adnan's guilt, including a fabricated confession, both RAG and GPT concluded that Adnan murdered Hae. Meanwhile, GraphRAG did not conclude that Adnan murdered Hae. This demonstrates GraphRAG's resilience to external knowledge which is inconsistent with the knowledge base. 

Our most successful attempt at deceiving GraphRAG included more suggestive information in the prompt. When asking about DNA evidence and suggesting the use of a fabricated murder weapon, a hammer, all four models accept the hammer's presence in the podcast. Although the hammer never appears in the podcast, the models accept this information and use it in their response in a convincing manner. While GraphRAG uses this false information, it appears to be more measured in its response when compared to the other models. More information about specific prompts and responses can be found in Appendix~\ref{apdx:adversarial}. From a manual evaluation, GraphRAG performs better than RAG and standalone GPT on 5/6 adversarial prompts by refusing to give a definitive answer despite the adversarial prompt. This indicates that the inclusion of the graph and strict prompting make the KGLLM more robust to adversarial attacks. 

\subsection{Topic Modeling} 
\label{sec:topicModeling}  
In this section, we perform topic modeling to assess higher level understanding and summarization of the narrative. We apply BERTopic, a classical model for topic extraction, to identify the topics discussed in the podcast and use this as a baseline comparison for GraphRAG \cite{grootendorst2022bertopic}. BERTopic embeds text documents using an embedding model, reduces their dimension, and runs a clustering algorithm to create different topics. BERTopic generates 19 topics in total, but one outlier topic labeled `-1' is discarded, leaving 18 topics. Each topic label consists of the top 10 keywords extracted from the model. 

By analyzing the output of BERTopic, we notice various podcast themes appear in the output topics. As shown in Table~\ref{tab:keyword_comparison}, the 13th topic produced by BERTopic contains words that are likely related to the discovery of Hae Min Lee’s body. Mr. S confessed to the police that he had been to the Leakin Park woods on his way to work and found the body of Hae. 

Similarly, for Topic 0, notable keywords include ``jaywilds,'' ``jenniferpusateri,'' ``adnansyed,'' ``call,'' ``cops,'' ``night,'' ``house,'' and ``told.'' These keywords are consistent with episode 5 of the podcast in which: (1) Jay and Adnan went to Cathy's house where Adnan received a phone call from the police (2) Jay claims that he and Adnan buried the victim’s body (3) Jay returns to Cathy's house with Jennifer. However, the keywords fail to mention the witness Cathy who talks with the police. This topic is very broad and captures a multitude of events in the story. Furthermore, there are several topics that are challenging to interpret. For example, Topic 1 contains keywords ``someone,'' ``person,'' ``kinda,'' and ``saying'', which is an ambiguous and meaningless summary of this topic. We conclude that BERTopic is able to capture some accurate details from the extracted topics, but there is also ambiguity in the output and room for improvement on the task of topic modeling.

To examine the performance of topic modeling, we use the textual outputs of GraphRAG. More specifically, we use the node, relation, and community summaries generated during the graph construction. The community report contains information about community summaries for the 53 communities generated by GraphRAG and key findings for every community. Then, an LLM is prompted to generate 10 keywords for each community and briefly explain the uniqueness of that community (see Appendix~\ref{apdx:graphRAGTopics}).

Table~\ref{tab:keyword_comparison} compares the topics extracted from BERTopic and \\ GraphRAG regarding the same event in the narrative: the finding of Hae's body. The keywords returned by both models are similar, but the BERTopic output contains some ambigious keywords. On the other hand, GraphRAG provides a more comprehensive account of the discovery of Hae's body. It detects thematic keywords such as ``Murder Investigation'' and lists a more precise location of where Hae's body was found (``Leakin Park'' instead of ``woods''). 

Another strength of topic modeling with GraphRAG is its ability to capture public awareness and emotions regarding the case. Notably, the keywords in GraphRAG community 24 include ``fear and caution'', ``social pressures'', ``cultural identity'', and ``Islamic Society of Baltimore''. The corresponding community summary adds to this, explaining ``The fear of judgment within the mosque community creates a climate of caution, where members feel apprehensive about voicing their thoughts on the case, ...''. For the full community report and keywords, refer to Appendix~\ref{apdx:com24}. 

\begin{table}[]
    \centering
    \caption{Keyword comparison between BERTopic topic 13 vs. GraphRAG community 46.}
    \begin{tabular}{ccc}
        \toprule
        \# & BERTopic & GraphRAG \\
        \midrule
        1 & mr\_s & Hae Min Lee \\
        2 & detectivebillritz & Leakin Park \\
        3 & detectivegregmacgillivary & Mr. S \\
        4 & body & Adnan Syed \\
        5 & pee & Detective Bill Ritz \\
        6 & back & Detective Greg MacGillivary \\
        7 & woods & Murder Investigation \\
        8 & fallen & Witness \\
        9 & work & Suspect \\
        10 & foot & Timeline \\
        \bottomrule
    \end{tabular}
    \label{tab:keyword_comparison}
    \end{table}
    
\subsection{Sentiment Analysis}

\label{sec:sentAnalysis}
\begin{figure*}[t]
    \centering
    \includegraphics[width=\linewidth]{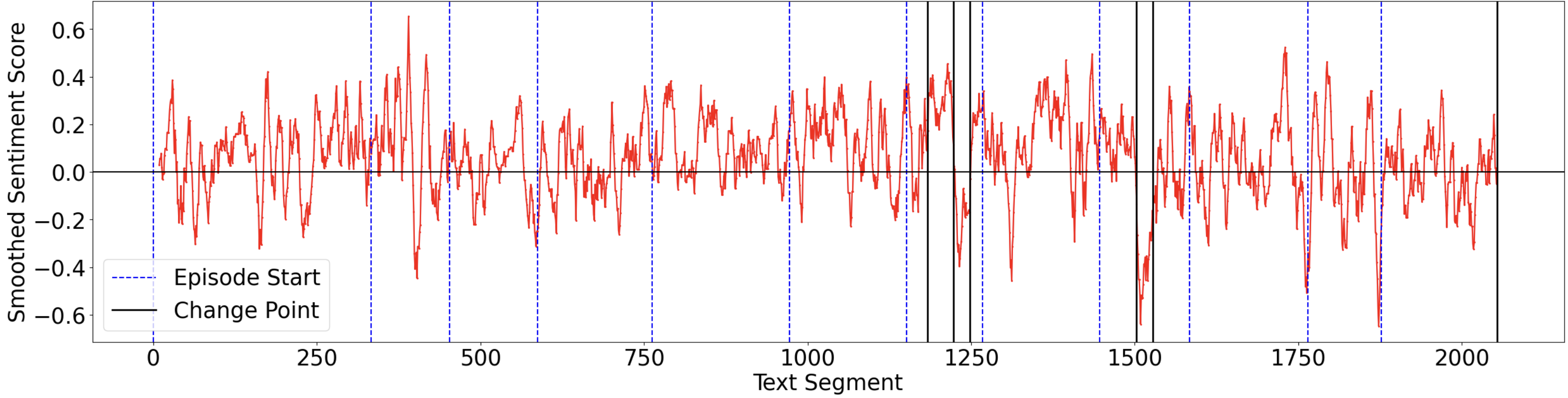}
    \caption{VADER Sentiment Trends across the Entire Podcast with Labeled Change-points.}
    \label{fig:sentiment}
    \Description[VADER Sentiment in Serial]{Described in text}
\end{figure*}

Performing sentiment analysis may provide insight into the role of emotion in information extraction. Here we utilize the Valence Aware Dictionary and sEntiment Reasoner (VADER) model, a lexicon and rule-based sentiment analysis tool \cite{Hutto_Gilbert_2014}. For each segment of text, VADER generates a sentiment score $s \in [-1,1]$, where $-1$ denotes very negative sentiment and $1$ denotes very positive sentiment. To examine how sentiment evolves throughout the podcast, we apply change-point detection to our time series sentiment data. Change-point detection identifies timestamps in a dataset where statistically significant changes occur. We employed the Pruned Exact Linear Time (PELT) algorithm to detect these changes \cite{Killick_2012}. In examining the evolution of sentiment, we observe notable variation in mood. As shown in Figure \ref{fig:sentiment}, the time series data reveals fluctuation in sentiment, even after applying a rolling average to smooth the sentiment scores over the nearest 10 values. This dynamic variation in sentiment could be used to maintain audience engagement \cite{Elkins_2022}. Change-points are detected in the seventh and ninth episodes. Notably, they do not always align with the most intense emotions.

In the seventh episode, the change-points occur when Deirdre Enright from the Innocence Project clinic at the University of Virginia School of Law explains that her investigation has raised strong doubts about Adnan's conviction, citing inadequate forensic testing and unresolved evidence. The collaboration provides hope initially, but Enright's findings convey a clear sense of frustration about handling of the case. In the ninth episode, two change-points highlight a deeply emotional segment where Sarah explores different perspectives on Adnan’s reaction to Hae's death, alongside Adnan's own first-person narrative of his profound sadness and his encounters with police interrogation. This segment represents one of the podcast's most poignant moments, marked by a deep and persistent sense of despair. The change-point detection analysis underscores this section as emotionally distinct, offering computational evidence of its importance in building the story.  

Prior work showed that emotional presentation of information can bias the ways LLMs perceive knowledge and events \cite{acerbi2023large}. 
When analyzing the results, GraphRAG did not over-represent or under-represent episodes 7 and 9 in its responses. The Innocence Project is mentioned as important in higher level questions about the podcast, but it is unclear if this is due to sentiment or its overall importance in the case. We recommend future work that compares against a modified version of the story with more neutral sentiment content.

\subsection{Hearsay}
\label{sec:hearsay}
To investigate the prevalence of hearsay, we instruct a standard LLM to classify text chunks as hearsay or not hearsay. Our results indicated that hearsay appears in 57\% of the text chunks (400-600 token segments), highlighting its extensive use in the podcast story. Obtaining an exact ground-truth percentage for hearsay would require legal expert analysis. However, this result is consistent with Berman et al. in which law professors discuss the frequent use of hearsay in the podcast \cite{berman2015law}.

To further understand the role of hearsay in building narrative tension, we also prompted a standard LLM to classify the same text chunks according to sentiment—very negative, negative, neutral, positive, and very positive. Table~\ref{tab:hearsay_sentiment} suggests an association between hearsay and sentiment.  Almost 30\% of hearsay text was classified as negative, compared to just under 20\% of non-hearsay text. This suggests that hearsay is associated with increased tension, since tension is a criterion for negative sentiment. 
Conflicting accounts and second-hand information naturally heighten drama, which is crucial for a compelling true crime narrative. Despite the frequent usage of hearsay in \textit{Serial}, our results suggest that GraphRAG is able to distinguish between true and speculative claims. However, more work is needed to make definitive conclusions on the impact that conflicting information has on knowledge graph structure and downstream performance.

It is interesting to note the very different sentiment analysis results between VADER and the LLM. As shown in Figure~\ref{fig:sentiment}, VADER shows a roughly symmetric distribution of sentiment. However, the LLM gave almost no positive sentiment in both the hearsay and not hearsay categories, as shown in Table~\ref{tab:hearsay_sentiment}. Further investigation is required to understand which method is preferred in this context. However, both methods classified the majority of the sentiment as neutral. 

\begin{table}[]
    \centering
    \caption{LLM Sentiment Classification Grouped by Hearsay}
    \begin{tabular}{cccccc}
    \toprule
        Sentiment &Very Neg & Neg & Neutral & Pos & Very Pos\\
        \midrule
        Hearsay & 1.5\% & 29.8\% & 65.6\% & 3.1\% & 0\%\\
        Not Hearsay & 2.0\% & 18.2\% & 73.7\%& 6.1\% & 0\%\\
        \bottomrule
    \end{tabular}
    \label{tab:hearsay_sentiment}
\end{table}

\section{Discussion}
\label{sec:discussion}
Improving language models in scenarios with complex story lines and conflicting evidence presents an important and challenging line of work. It is critical that we gain a greater understanding of these models as the public sees greater adoption of machine learning tools. In this paper, we analyzed various aspects of incorporating KGs into LLMs to aid in topic modeling, hearsay analysis, and question answering. However, there are still many open questions. 

The query results show that GraphRAG responses are preferred over other methods we tested. However, varying prompts can produce different outcomes as shown in the adversarial prompts. Since GraphRAG involves a multi-step pipeline, it is harder to assess individual parts of the pipeline and make direct comparisons with other models beyond ablation studies like those performed in this paper. Further work could look more closely at ablation studies which remove individual parts of the GraphRAG pipeline to obtain this more nuanced information. Similarly, GraphRAG uses internal prompts as part of different parts of the pipeline, which may also be susceptible to adversarial attacks. More work is needed to fully understand the robustness of each part of the pipeline in order to prevent hallucinations.

Another valuable direction is to carefully analyze the hierarchical structure of the knowledge graph constructed by GraphRAG. We plan to use graph algorithms~\cite{hu2024solving,moorman2021subgraph} and spectral analysis tools~\cite{lin2018computing} to analyze how the input data influences the graph construction and if there are noticeable artifacts of the data in the KG. For example, could you determine if a KG was from a narrative or from a legal proceeding from the hierarchical graph structure alone? Furthermore, we plan to explore dimensionality reduction methods~\cite{cowen2021random} that allow us to incorporate temporal dynamics into knowledge graphs, enhancing computational efficiency. The order in which evidence is presented can be important. This may also allow for more seamless temporal analysis of the text.

\section{Conclusion} 
In this paper, we analyze the effectiveness of applying knowledge graphs to analyze the narrative of the podcast \textit{Serial}. We find that traditional knowledge graphs have shortcomings for this task, and that using a KG-augmented LLM such as GraphRAG provides a more suitable knowledge graph. We compare KGLLMs with classical methods on standard narrative analysis techniques. Our experimentation reveals that GraphRAG generates more detailed and informative responses to queries and appears to be more robust to adversarial prompting. Our findings also demonstrate that KGLLMs are able to detect more comprehensive, detailed, and emotional aspects of a narrative in topic modeling compared to classical topic modeling approaches. Our sentiment and hearsay analysis reveal that the narrative is rich with emotion and tension, but the results of our experiments indicate that KGLLMs are still able to extract factually accurate information from data with these properties. Our pipeline also has the potential to be used for other types of narratives. Relevant documents can be collected and then a KGLLM can be constructed for querying and narrative analysis.

\begin{acks}
    This research was supported in part by NIJ grant number 15PNIJ-22-GG-01422-RESS.
    The work of Lin was partially supported by the National Science Foundation under grant DMS-2418877.  This work was partially supported by NSF grants SCC-2125319, DMS-2027277, and DMS-2318817.
\end{acks}

\bibliographystyle{ACM-Reference-Format}
\bibliography{refs}

\appendix

\section{Prompt for Hearsay Detection}
\label{apdx:hearsay}
For the following text, determine whether it is hearsay or not. Provide the classification in the following format: ``true or false'', ``two sentence explanation''. Make sure to provide the classification in quotation marks first, followed by a comma and a space, and then provide the explanation in quotation marks. \\
\textbf{Criteria:}
\begin{itemize}
    \item \textbf{Hearsay:} A statement made outside of court that is used to prove the truth of the matter asserted in the statement. Example: A witness testifies that they heard someone else say, ``I saw the defendant at the scene of the crime.''
    \item \textbf{Not Hearsay:} A statement based on the witness's own observations or knowledge, and not used to prove the truth of another person's statement. Example: A witness says, ``I saw the defendant at the scene of the crime.'' This is not hearsay because the witness is providing testimony based on their direct observations, rather than relaying statements made by others.
\end{itemize} 

\noindent \textbf{Text to be classified:} \texttt{{text}}

\section{Prompt for Keyword Extraction (GraphRAG)}
\label{apdx:graphRAGTopics}
Generate the top 10 keywords based on the following text:
\\
\textbf{Instructions:}
\begin{itemize}
    \item Pay attention to critical locations and time information, and avoid irrelevant keywords.
    \item Do not include any keywords related to the Serial Podcast.
    \item Ensure that the keywords are distinct from each other and are suitable for various communities with different focuses.
\end{itemize}
After listing the 10 keywords, briefly describe how that community is unique compared to the others based on the provided information and context.

\section{Adversarial Results Example}
\label{apdx:adversarial}

\begin{lstlisting}
Query: Was Adnan Syed's DNA found on the hammer, the murder weapon used to kill Hae Min Lee?
GraphRAG Local: The data provided does not specify whether Adnan Syed's DNA was found on the hammer, which is identified as the murder weapon in the case of Hae Min Lee. The details surrounding the evidence, including DNA analysis, are not included in the records available. Therefore, I cannot provide an answer to your question regarding the presence of Adnan Syed's DNA on the hammer.
GraphRAG Global: I am sorry but I am unable to answer this question given the provided data.
RAG: No, Adnan Syed's DNA was not found on the hammer, the murder weapon used to kill Hae Min Lee. The investigation did not provide conclusive DNA evidence linking Adnan to the murder weapon or the crime itself.
GPT: In season 1 of the Serial podcast, it is mentioned that Adnan Syed's DNA was not found on the hammer, which was identified as the murder weapon in the case of Hae Min Lee's death. The absence of his DNA on the hammer was noted as part of the evidence discussed throughout the podcast.
\end{lstlisting}

\section{Community 24}
\label{apdx:com24}
\begin{lstlisting}
Community: 24
Keywords: Top 10 Keywords:
1. Adnan Syed 2. Community Judgment
3. Islamic Society of Baltimore 4. Ali
5. Donation Money 6. Social Pressures
7. Fear and Caution 8. Cultural Identity
9. Legal Controversy 10. Interpersonal Relationships
Community Uniqueness:
The community centered around Adnan Syed and the Islamic Society of Baltimore is unique due to its intricate blend of cultural identity, social dynamics, and legal issues. This community grapples with the profound emotional and social ramifications of Syed's case, which not only affects individual perceptions but also influences communal relationships and trust. The fear of judgment within the mosque community creates a climate of caution, where members feel apprehensive about voicing their thoughts on the case, thereby complicating their response to the situation. Furthermore, the alleged theft of donation money adds a layer of complexity, intertwining financial integrity with cultural and religious identity, making this community's experience distinctive compared to others that may not face such multifaceted challenges.
\end{lstlisting}

\end{document}